\documentclass{article}
\usepackage{spconf,amsmath,graphicx}
\usepackage{booktabs}
\usepackage{xurl}
\usepackage{enumitem}
\usepackage{amsmath,amssymb,amsfonts}
\usepackage{algorithmic}
\usepackage{graphicx}
\usepackage{textcomp}
\usepackage{xcolor}
\usepackage{comment}
\usepackage{booktabs}
\usepackage{multirow}
\usepackage{siunitx}
\usepackage[table]{xcolor} 

\usepackage[numbers,sort&compress]{natbib}
\setlist{nosep, leftmargin=14pt}

\usepackage{mwe} 

\title{Hospital-Specific Bias in Patch-Based Pathology Models}
\name{Mengliang Zhang}
\address{
  CSE Department, The University of Texas at Arlington
}

\begin{document}
%
\maketitle
\begin{abstract}
Pathology foundation models (PFMs) achieve strong performance on diverse histopathology tasks, but their sensitivity to hospital-specific domain shifts remains underexplored. We systematically evaluate state-of-the-art PFMs on TCGA patch-level datasets and introduce a lightweight adversarial adaptor to remove hospital-related domain information from latent representations. Experiments show that, while disease classification accuracy is largely maintained, the adaptor effectively reduces hospital-specific bias, as confirmed by t-SNE visualizations. Our study establishes a benchmark for assessing cross-hospital robustness in PFMs and provides a practical strategy for enhancing generalization under heterogeneous clinical settings. Our code is available at \url{https://github.com/MengRes/pfm_domain_bias}.
\end{abstract}
\section{Introduction}
\label{sec:intro}

Recent advances in pathology foundation models (PFMs) have enabled powerful and transferable representations from large-scale histopathology datasets, achieving strong performance in tasks such as cancer grading~\cite{vorontsovFoundationModelClinicalGrade2024}, subtyping~\cite{heBoostingPathologyFoundation2025}, and lesion detection~\cite{xiangVisionLanguageFoundation2025a}. However, these models remain sensitive to staining variations and domain shifts across institutions. Variations in staining protocols, scanners, and tissue preparation introduce strong hospital-specific biases, which can degrade model performance when applied to data from unseen hospitals. Traditional stain normalization methods, such as Macenko~\cite{m.macenkoMethodNormalizingHistology2009} and Reinhard~\cite{e.reinhardColorTransferImages2001}, partially reduce color differences but cannot remove domain-specific cues embedded in deeper feature representations.

\begin{figure}[htbp]
    \centering
    \includegraphics[width=0.6\columnwidth]{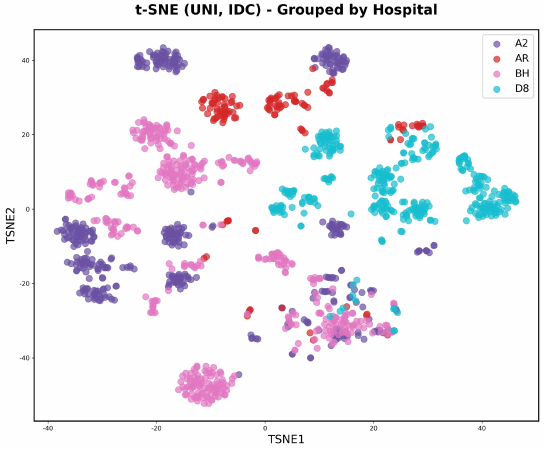}
    \caption{t-SNE of patch features for IDC in TCGA-BRCA, extracted by the UNI model. Different color represent different hospital source, showing hospital-specific clustering.}
    \label{fig:tsne_example}
\end{figure}

In fact, PFMs often encode hospital-specific information alongside disease-relevant features. Figure~\ref{fig:tsne_example} illustrates this effect using t-SNE visualization of patch-level embeddings: points are clearly grouped by hospital origin, indicating that even high-performing PFMs can leak institution-specific signals into their latent space. Several prior studies also investigated domain bias in pathology foundation models (PFMs). Even with stain normalization~\cite{m.macenkoMethodNormalizingHistology2009}, embeddings often retain hospital-specific information, which can degrade performance on unseen institutions and limit clinical generalization. Existing robustness metrics and alignment methods either fail to assess dataset-level bias or do not explicitly preserve disease-discriminative features~\cite{komenRobustFoundationModels2025}. This motivates the need for methods that explicitly reduce hospital-domain influence while preserving disease-related features. While foundational methods for adversarial domain adaptation, such as the Domain-Adversarial Neural Network (DANN)~\cite{ganin2016domain}, have demonstrated success in other areas, their application as a lightweight, plug-and-play adaptor for large-scale, pre-trained PFMs remains underexplored. Our work aims to bridge this gap by systematically evaluating and applying this technique to frozen PFM backbones.

\begin{figure*}[htbp]
    \centering
    \includegraphics[width=0.8\textwidth]{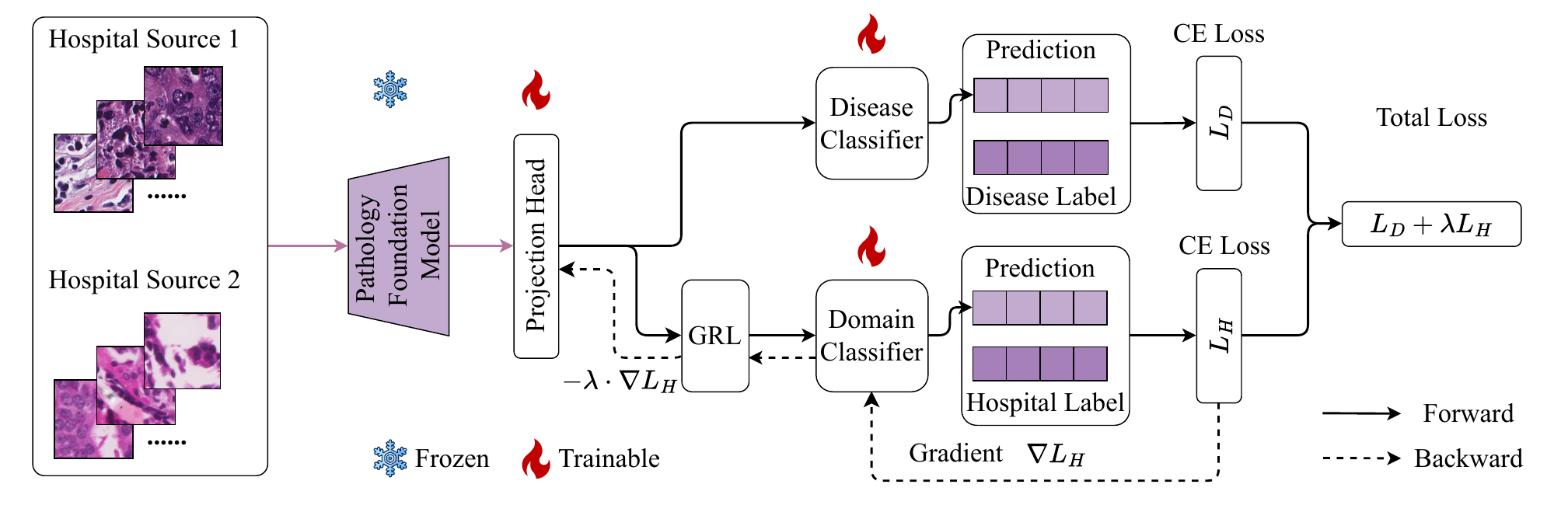}
    \caption{Adversarial framework for pathology classification: features from a shared foundation model are optimized for disease prediction while suppressing hospital-specific cues via a gradient reversal layer (GRL).}
    \label{fig:adversarial_method}
\end{figure*}

In this work, we systematically evaluate hospital-specific bias in state-of-the-art pathology foundation models (PFMs) using patch-level embeddings from the TCGA~\cite{TCGA_BRCA_WSI_ISBI} dataset. Through t-SNE visualization, we observe that even high-performing PFMs, such as Phikon~\cite{filiot2023phikon} and UNI~\cite{chenGeneralPurposeFoundationModel2024}, encode hospital-specific information alongside disease-relevant features, which may compromise generalization to unseen institutions. To mitigate this effect, we propose a lightweight adversarial adaptor module that can be appended to pretrained PFMs without retraining the backbone. The adaptor suppresses hospital-domain signals in the latent space while largely preserving disease-discriminative representations.

Our experiments across multiple hospitals demonstrate that the proposed framework effectively reduces hospital-specific influence on feature representations, as confirmed by lower hospital classification accuracy. Meanwhile, disease classification performance is maintained or slightly improved. Overall, this study highlights a previously underexplored limitation of current PFMs—their sensitivity to institutional domain—and provides a practical, plug-and-play approach to enhance their robustness under heterogeneous staining and cross-hospital settings. We anticipate that our evaluation pipeline and adaptor module will serve as a foundation for future research on domain-invariant representation learning in computational pathology.

\section{Method}
\label{sec:method}

\subsection{Patch Selection}
\vspace{-0.2cm}
We construct a multi-hospital WSI cohort with balanced disease classes and hospital sources. Tissue regions are segmented using CLAM \cite{luDataEfficientWeaklySupervised2021}, and fixed-size patches are extracted via a sliding-window strategy. To reduce patch-level label noise inherent to WSI-level supervision, we apply quality control followed by CONCH \cite{luVisualLanguageFoundationModel2024} zero-shot filtering, retaining only patches whose top-1 prediction matches the WSI label with confidence above 0.8. The selected patches are used to train two independent baseline MLP classifiers for disease and hospital-source prediction. The classifiers do not share weights and are trained separately to assess the discriminative quality and domain leakage of frozen pathology foundation model features. Each MLP consists of two hidden layers (512 and 256 units) with ReLU activations. High hospital-source accuracy indicates stronger domain bias, whereas high disease accuracy reflects preserved diagnostic information.


\vspace{-0.3cm}
\subsection{Adversarial Method}
\vspace{-0.2cm}

Inspired by the principles of Domain-Adversarial Neural Networks (DANN) ~\cite{ganin2016domain}, we introduce a lightweight adversarial framework to suppress hospital-specific cues while preserving disease information in Figure\ref{fig:adversarial_method}. The framework includes a frozen encoder $E(\cdot)$, a trainable projection head $A(\cdot)$, a disease classifier $C_y(\cdot)$, and a domain classifier $C_d(\cdot)$ connected via a Gradient Reversal Layer (GRL). Let the dataset be $\mathcal{D} = \{(x_i, y_i, d_i)\}_{i=1}^N$, with patch $x_i$, disease label $y_i$, and hospital label $d_i$. The encoder produces features $\mathbf{f}_i = E(x_i)$, and the projection head maps them to

\begin{equation}
\mathbf{z}_i = A(\mathbf{f}_i) \in \mathbb{R}^{D'}
\end{equation}
Disease and domain predictions are
\begin{equation}
\hat{y}_i = C_y(\mathbf{z}_i), \quad
\hat{d}_i = C_d(\mathrm{GRL}(\mathbf{z}_i))
\end{equation}

Gradient Reversal Layer (GRL) acts as identity in forward pass and reverses gradients in backward pass:
\begin{equation}
\frac{\partial \mathcal{L}}{\partial \mathbf{z}_i} \leftarrow -\lambda \frac{\partial \mathcal{L}}{\partial \mathbf{z}_i}
\end{equation}
encouraging $\mathbf{z}_i$ to retain disease-discriminative information while removing hospital-specific cues.

The total loss is
\begin{equation}
\mathcal{L}_\text{total} = \mathcal{L}_\text{D} + \lambda \mathcal{L}_\text{H}
\end{equation}
with
\begin{align}
\mathcal{L}_\text{D} &= \frac{1}{N} \sum_{i=1}^N \mathrm{CE}(C_y(\mathbf{z}_i), y_i) \\
\mathcal{L}_\text{H} &= \frac{1}{N} \sum_{i=1}^N \mathrm{CE}(C_d(\mathrm{GRL}(\mathbf{z}_i)), d_i)
\end{align}
Only the projection head and classifiers are trained; the encoder is frozen. At inference, GRL and domain classifier are discarded: 
\begin{equation}
\hat{y}_i = C_y(\mathbf{z}_i)
\end{equation}

In Figure~\ref{fig:adversarial_method}, we train two independent MLP classifiers on the extracted patches: one for disease classification and one for hospital-source identification. The classifiers do not share weights and are trained separately to assess the discriminative capacity of the frozen backbone features for each task. Performance is evaluated using accuracy, AUC, and F1 score. High accuracy in hospital-source classification indicates strong domain bias, whereas high accuracy in disease classification reflects better disease-discriminative representations.

\section{Experiment}
\label{sec:experiment}

We focus on whole-slide image (WSI) samples from the TCGA-BRCA dataset~\cite{TCGA_BRCA_WSI_ISBI}.
We identify the four hospitals contributing the largest numbers of WSIs and randomly select 20 WSIs per hospital, restricted to patients who are White, female, and aged 60–79 years.
From each selected WSI, 500 image patches of size $256 \times 256$ at 40$\times$ magnification are uniformly sampled. Since randomly sampled patches may not always reflect the WSI-level diagnosis (e.g., an IDC-labeled WSI may contain normal tissue), we filter patches using the CONCH foundation model~\cite{luVisualLanguageFoundationModel2024}.
Only patches whose predicted label matches the WSI label and whose confidence exceeds 0.8 are retained, resulting in 4,029 high-confidence patches across two disease categories: Invasive Ductal Carcinoma (IDC) and Invasive Lobular Carcinoma (ILC).

The filtered patches are used to extract features from representative pathology foundation models, including ResNet-50~\cite{he2016deep}, Giga-Path~\cite{xu2024gigapath}, UNI~\cite{chenGeneralPurposeFoundationModel2024}, UNI2-H~\cite{chenGeneralPurposeFoundationModel2024}, CONCH~\cite{luVisualLanguageFoundationModel2024}, TITAN~\cite{dingMultimodalWholeSlide2024}, MUSK~\cite{xiangVisionLanguageFoundation2025a}, H-Optimus-0~\cite{hoptimus0}, Phikon~\cite{filiot2023phikon}, Phikon-v2~\cite{filiotPhikonV2LargePublic2024}, and Virchow~\cite{vorontsovFoundationModelClinicalGrade2024}.
Patch-level features are visualized with t-SNE to inspect hospital-specific clustering and domain bias.

For adversarial training, we use a batch size of 64 over 30 epochs, set the adversarial strength $\lambda$ to 0.5, and employ a projection head with hidden dimension 512. The value of $\lambda=0.5$ was determined empirically by a parameter sweep on a held-out validation fold, selected to best balance the reduction in hospital-source AUC while maintaining high disease classification ACC.
Five-fold cross-validation ensures that patches from the same WSI appear in only one fold, preventing information leakage across training and validation sets.

\begin{table}[htbp]
\centering
\caption{Performance of different models on hospital-source (domain) and disease classification using MLP and our method. CONCH is shown in gray as reference (patches extracted from it). Best results for MLP and our method are highlighted in green, worst in red, respectively. Mean $\pm$ std is reported for accuracy (ACC) and AUC.}
\label{tab:combined_results_best_worst_2dec_full_compact}
\scriptsize
\renewcommand{\arraystretch}{0.9}  
\setlength{\tabcolsep}{3pt}        
\begin{tabular}{l l c c c c}
\toprule
\textbf{Model} & \textbf{Method} & \textbf{Hosp ACC} & \textbf{Hosp AUC} & \textbf{Disease ACC} & \textbf{Disease AUC} \\
\midrule
\multirow[0]{2}{*}{CONCH}     
  & MLP  & \cellcolor{gray!20}0.61$\pm$0.06 & \cellcolor{gray!20}0.84$\pm$0.03 & \cellcolor{gray!20}1.00$\pm$0.00 & \cellcolor{gray!20}1.00$\pm$0.00 \\
  & Ours & \cellcolor{gray!30}0.23$\pm$0.17 & \cellcolor{gray!30}0.59$\pm$0.13 & \cellcolor{gray!30}1.00$\pm$0.00 & \cellcolor{gray!30}1.00$\pm$0.00 \\
\midrule
\multirow[0]{2}{*}{GIGA\_PATH}     
  & MLP  & 0.68$\pm$0.12 & 0.91$\pm$0.04 & 0.92$\pm$0.03 & 0.98$\pm$0.01 \\
  & Ours & 0.24$\pm$0.19 & 0.57$\pm$0.11 & 0.92$\pm$0.02 & 0.96$\pm$0.01 \\
\midrule
\multirow[0]{2}{*}{H\_OPTIMUS} 
  & MLP  & 0.81$\pm$0.09 & 0.97$\pm$0.02 & 0.91$\pm$0.05 & 0.97$\pm$0.02 \\
  & Ours & 0.34$\pm$0.16 & 0.67$\pm$0.13 & 0.92$\pm$0.04 & 0.96$\pm$0.03 \\
\midrule
\multirow[0]{2}{*}{MUSK}      
  & MLP  & 0.69$\pm$0.09 & 0.90$\pm$0.05 & \cellcolor{green!30}0.94$\pm$0.01 & \cellcolor{green!30}0.99$\pm$0.00 \\
  & Ours & 0.29$\pm$0.16 & 0.50$\pm$0.00 & \cellcolor{green!30}0.94$\pm$0.01 & \cellcolor{green!30}0.98$\pm$0.01 \\
\midrule
\multirow[0]{2}{*}{PHIKON}    
  & MLP  & 0.85$\pm$0.07 & 0.98$\pm$0.01 & 0.85$\pm$0.09 & 0.94$\pm$0.04 \\
  & Ours & 0.45$\pm$0.12 & 0.74$\pm$0.11 & 0.88$\pm$0.09 & 0.93$\pm$0.06 \\
\midrule
\multirow[0]{2}{*}{PHIKON-V2} 
  & MLP  & \cellcolor{red!30}0.92$\pm$0.03 & \cellcolor{red!30}0.99$\pm$0.00 & 0.87$\pm$0.08 & 0.95$\pm$0.04 \\
  & Ours & \cellcolor{red!30}0.48$\pm$0.19 & \cellcolor{red!30}0.76$\pm$0.17 & 0.86$\pm$0.09 & 0.94$\pm$0.04 \\
\midrule
\multirow[0]{2}{*}{RESNET50}  
  & MLP  & \cellcolor{green!30}0.60$\pm$0.09 & \cellcolor{green!30}0.83$\pm$0.05 & \cellcolor{red!30}0.84$\pm$0.03 & \cellcolor{red!30}0.93$\pm$0.02 \\
  & Ours & 0.25$\pm$0.16 & 0.51$\pm$0.03 & \cellcolor{red!30}0.86$\pm$0.05 & \cellcolor{red!30}0.93$\pm$0.03 \\
\midrule
\multirow[0]{2}{*}{TITAN}     
  & MLP  & 0.57$\pm$0.07 & 0.82$\pm$0.04 & 0.94$\pm$0.01 & 0.98$\pm$0.01 \\
  & Ours & 0.23$\pm$0.11 & 0.55$\pm$0.09 & 0.93$\pm$0.02 & 0.98$\pm$0.01 \\
\midrule
\multirow[0]{2}{*}{UNI}       
  & MLP  & 0.78$\pm$0.07 & 0.95$\pm$0.03 & 0.93$\pm$0.03 & 0.98$\pm$0.01 \\
  & Ours & 0.22$\pm$0.11 & 0.58$\pm$0.06 & 0.93$\pm$0.04 & 0.96$\pm$0.04 \\
\midrule
\multirow[0]{2}{*}{UNI2-H}    
  & MLP  & 0.79$\pm$0.10 & 0.96$\pm$0.02 & 0.92$\pm$0.04 & 0.98$\pm$0.01 \\
  & Ours & 0.19$\pm$0.11 & 0.51$\pm$0.05 & 0.92$\pm$0.04 & 0.97$\pm$0.02 \\
\midrule
\multirow[0]{2}{*}{VIRCHOW}   
  & MLP  & 0.71$\pm$0.15 & 0.91$\pm$0.07 & 0.91$\pm$0.05 & 0.98$\pm$0.01 \\
  & Ours & \cellcolor{green!30}0.18$\pm$0.11 & \cellcolor{green!30}0.49$\pm$0.03 & 0.91$\pm$0.05 & 0.97$\pm$0.01 \\
\bottomrule
\end{tabular}
\end{table}

\begin{figure}[htbp]
    \centering
    \includegraphics[width=1.0\columnwidth]{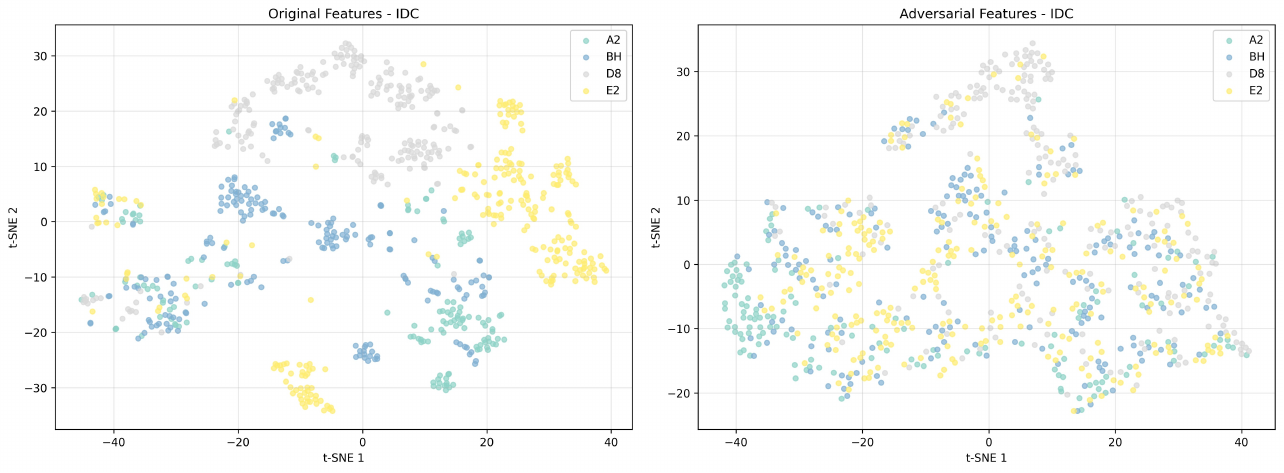}
    \caption{t-SNE clustering results of the original patch features using the UNI model (left) and the t-SNE clustering results of the features after adversarial training (right). All patches share the same label, while points in different colors correspond to different hospitals.}
    \label{fig:tsne_uni}
\end{figure}

\vspace{-0.3cm}
\subsection{Results}

\vspace{-0.2cm}
We conducted experiments on the TCGA-BRCA~\cite{TCGA_BRCA_WSI_ISBI} dataset to evaluate pathology foundation models from multiple perspectives, including t-SNE feature visualization, disease classification performance, and hospital-source classification performance. Table \ref{tab:combined_results_best_worst_2dec_full_compact} reports the performance of multiple pathology foundation models on hospital-source (domain) and disease classification tasks, evaluated using accuracy (ACC) and Area Under the Curve (AUC). For each model, we compare the baseline MLP method and our proposed adversarial approach. The CONCH model, shown in gray, serves as a reference baseline.

Overall, our method consistently reduces hospital-specific bias while maintaining or improving disease classification performance. This is demonstrated across most models by a significant reduction in hospital-source ACC and AUC. For instance, under the baseline MLP, both UNI and UNI2-H exhibited high domain bias, with hospital AUCs of $0.96 \pm 0.02$. After applying our method, their hospital AUCs dropped significantly to $0.51 \pm 0.05$. Similarly, Virchow's hospital AUC decreased from $0.91 \pm 0.07$ to $0.49 \pm 0.03$.

While effectively reducing domain bias, our method successfully preserves high disease classification performance. Notably, the MUSK model showed the highest baseline disease ACC ($0.94 \pm 0.01$), and our method maintained this performance at $0.93 \pm 0.02$. Models like UNI and UNI2-H saw their disease ACC remain almost unchanged ($0.93 \pm 0.03$ and $0.92 \pm 0.04$, respectively). In contrast, Phikon-V2, which exhibited one of the highest baseline domain biases (Hosp AUC $0.99 \pm 0.00$), saw its bias reduced (Hosp AUC $0.76 \pm 0.17$) while its disease ACC was maintained ($0.87 \pm 0.08$ vs $0.86 \pm 0.09$).

To illustrate the quantitative results in Table \ref{tab:combined_results_best_worst_2dec_full_compact}, 
Figure \ref{fig:tsne_uni} shows t-SNE visualizations of UNI model features. The left panel displays features from the pre-trained model, forming clear clusters by hospital, indicating strong domain bias. The right panel shows features after applying our adversarial adaptor, where hospital-specific clustering is largely dissolved, 
demonstrating effective suppression of domain signals while preserving disease-related features. Overall, the results highlight that our approach achieves a desirable balance: reducing hospital/source information (domain-invariant) while preserving or enhancing disease classification performance.

\section{Conclusion}
\label{sec:conclusion}

In this work, we addressed the issue of domain bias in pathology foundation models. We demonstrated that latent hospital-specific features can significantly affect model performance across institutions. To mitigate this, we proposed a lightweight adversarial training framework that effectively removes such biases while preserving disease classification accuracy. Our experiments on TCGA datasets validate the effectiveness of this approach. These findings highlight the importance of explicitly modeling and eliminating hidden domain biases to develop robust, generalizable, and fair medical AI systems, providing a practical blueprint for future clinical deployment.

\bibliographystyle{IEEEbib}
\bibliography{strings,refs}

\end{document}